# A Swarm-Coordinated Multi-Robot System for Early Stress Detection in Agricultural Rows Using Multimodal Leaf Sensing

**Rishi Gupta[1], Astha Goyal[2*], and Vinay Vishwakarma[3]**

**[1]Delhi Public School Vasant Kunj, New Delhi, India, rishi.gupta2707@gmail.com**
**[2]*Corresponding author: Assistant Professor, University of Delhi, astha.goyal@keshav.du.ac.in**
**[3]Mentor, On My Own Technology Pvt. Ltd, Mumbai, India, vinay.vishwakarma@omotec.in**

## Abstract

**Early stress detection in crops is a necessity today to improve efficiency and reduce waste of time, money, and effort. However, most modern techniques, such as hyperspectral imaging and AI-based systems, are too costly and complex for medium and small-scale farmers to implement. This paper showcases CropSentry, a low-cost, ground-based multi-robot system that uses multimodal leaf sensing to continuously monitor crop health by tracking stress levels. The system comprises two autonomous bots that continuously detect leaf color and environmental data row by row. The observations are spatially mapped and sent over to the master bot, which uses color-coded row segments to generate a real-time web-based dashboard displaying crop health. After 63 observations were collected during the experiments, the results showed an overall crop health classification accuracy of 84.12%, with 82.60% for healthy plants, 88% for nutrient-deficient plants, and 80% for diseased plants. Also, 100% wireless communication success rate across 10 slave observations was achieved. Close-range leaf inspection across multiple bots can detect early stress in crops while remaining affordable, accessible, and scalable. It provides farmers with timely information to improve resource utilization and crop management.**

## I. Introduction

There is an increasing threat to agriculture from crop stressors, including drought, nutrient deficiencies, insects, and diseases. According to the FAO, pests and diseases account for up to 40% of annual global crop losses [1]. This costs farmers approximately $290 billion each year. Therefore, early detection of crop stress is critical to reducing yield losses, increasing resource efficiency, and helping ensure food security worldwide. The traditional method of monitoring crops involves manual inspection, which is laborious and time-consuming, and often makes it impossible to determine whether a problem exists until it is too late. More recently, technologies such as UAVs, hyperspectral imaging, and AI-based monitoring systems have been developed to enable early detection of crop stressors. Although these technologies are very accurate, they tend to have a high initial cost, typically require specialized equipment to operate, and, consequently, are often not feasible for small- or medium-sized farming operations.

This paper presents a swarm-coordinated multi-robotic system for early crop stress detection utilizing multimodal leaf sensing to overcome limitations. The system consists of two low-cost autonomous ground robots that are equipped with TCS34725 color sensors, environmental sensors, and wireless communication modules. The robots will travel through different rows of crops and will collect both color and environmental data from the leaves and then generate a spatial map representing the health of the crops. The major advantage of this solution, compared to aerial-based solutions, is that the leaf inspection will take place at close proximity, which allows for a more accurate measurement of the subtle differences in color that indicate the presence of crop stress. CropSentry also represents an affordable, scalable, and accessible solution for precision agriculture, providing farmers with timely information on how to address crop problems.

## II. Literature Review

The purpose of this paper by Minni et al. [2] is to provide an innovative implementation of image processing methods that can detect leaf nutrient deficiency in plants. The study specifically examines the differences in color between leaves suffering from nutrient deficiency, as well as utilizing color texture features to identify specific symptoms typical of leaf nutrient deficiency. The method develops a method to process digital photographs of plants and analyze the features extracted from them for early diagnosis of plant health.

Kohzuma et al. [3] in their research demonstrated an alternative, cost-effective, and non-destructive way for monitoring plant health, using digital images to obtain signals of early detection for farmers and allowing them to respond quickly and appropriately to potential problems with their crops before it is too late.

A low-cost RGB color sensor system has been developed to automate the sorting of fruits by determining the color of each fruit. Elwakeel et al. [4] designed and built an automated setup capable of sorting fruits with high precision. Experimental results show that this low-cost RGB color sensor system can accurately sort fruits with high precision while still being affordable. The research highlights the use of RGB sensing technology in agricultural automation as color-based classification improves sorting efficiency, decreases labor requirements, and improves the overall quality of produced goods.

Advanced technologies and data fusion techniques are reviewed in this paper by Ahmad et al. [5] for site-specific crop monitoring applications. The authors discuss how using a combination of sensors, imaging systems, drones, IoT devices, and machine learning algorithms has improved the decision-making process when it comes to agriculture. Overall, combining multiple sensing technologies can greatly improve precision farming by allowing for the efficient use of resources and better crop management plans.

This paper by Shi et al. [6] analyzes methods of navigation for agricultural robots and autonomous vehicles, with the focus being on navigating through row-crop fields. Authors analyze computer vision, sensor, and machine learning techniques that can be utilized to detect rows of crops, as well as aid robots in traversing the rows. The conclusion is that advanced row-detection systems are critical for enabling fully autonomous navigation and for reducing operational errors in order to increase efficiency within agriculture.

Porto and Arthur Jose Vieira [7] focused on the design and development of an agricultural mobile robot through its architectural characteristics. The research includes the robot's mechanical design, the robot's control devices and methods, the components used for sensing the environment, and how the robot will use those components to navigate in the environment. This research provides a systematic design methodology for designing and developing agricultural mobile robots that can operate effectively in dynamic environments and demonstrates how this work advances agricultural automation.

The researchers Fathima et al. [8] describe an embedded IoT-based real-time nutrient monitor used to monitor nitrogen, phosphorus, and potassium (NPK) in soil. This IoT-based monitor provides real-time assessment of the quality of soils, as well as enables the transmission of the data collected in real-time using wireless technologies via sensors and microcontroller technologies. The researchers have shown that IoT-enabled nutrient monitoring can provide the information necessary for making timely decisions when managing crops, for maximizing yield potential through optimizing the application of fertilizers, and for using less energy in applying excess fertilizer.

Shoaib et al. [9] present a comprehensive overview of new technologies used to detect plant stress through multimodal imaging and machine learning in this paper. Included in the paper are spectral imaging, thermal imaging, RGB imaging, and smartphone detection systems; ML models are assessed for their ability to find various plant stresses. The authors note that using multiple sensing methods together provides better accuracy in detecting plant stress and supports the development of simple-to-use, easy-to-implement field monitoring methods.

An overview of the current trends and uses of mobile robot systems in modern-day precision agriculture is discussed in this paper by Yépez-Ponce et al. [10]. Herein, the authors describe the types of robotic systems that are being used for planting, monitoring, harvesting, etc., as well as other precision agricultural operations. They also discuss the key technologies behind these robotic systems, including artificial intelligence, automatic navigation, and integrating sensors. It was concluded that as the use of mobile robots in agriculture continues to grow, this will increase both productivity and operational efficiency.

The paper by Alawode et al. [11] explores the use of artificial intelligence in the areas of climate-smart agriculture and fraud-proofing systems for financing green projects. The authors suggest AI-driven systems that can connect precision agriculture technologies with transparent financial tracking. The work also examines how intelligent monitoring systems could help achieve sustainability, optimize resource use in the agricultural industry, and increase output while maintaining accountability for financing activities.

The authors Velasquez et al. [12] present a reliable agricultural robot row-following system using multimodal sensor fusion. Combining data from different types of sensors (for example, GPS) will enable greater accuracy in robot navigation under different field conditions. This system was assessed for its ability to continue to maintain reliable tracking of its path while being subjected to external environmental perturbations and limitations associated with sensor types. It was determined that using multiple sensors improved the stability of the robot's operation and improved its performance when navigating through fields, thus indicating its potential utility for precision agriculture.

Walsh et al. [13] describe recent developments in imaging technology and artificial intelligence for detecting stress in crops through reviewing available literature. Some examples of such technologies include RGB cameras, multispectral sensors, thermal sensors, and deep learning algorithms. The authors explored emerging patterns as well as the future directions, barriers, and potentials associated with providing automated monitoring of crops. It was concluded that with the application of advanced imaging system technologies and AI methodologies, the detection of crop stress can be achieved, at scale, accurately, and without harming the crops, thus providing support for smart farming efforts.

The building of an autonomous, GPS navigational-controlled robot for precision agriculture purposes was presented in this study by Ünal et al. [14]. The work involved using GPS, wireless communications, and autonomous control systems within an automated robot to successfully execute agricultural tasks. Experimental tests were conducted to demonstrate successful robot navigation and task accomplishment. The results of this research have shown that by using a GPS automated robot, there is much potential for robots to be used in the agricultural sector with reduced man-hours requirements, improved task accuracy, and enhanced precision farming.

The purpose of this article by Li et al. [15] is to present in depth the use of deep learning approaches for detecting and identifying diseases in plants. The authors focus on the use of convolutional neural networks, transfer learning technology, and image-based diagnostic systems, which have been deployed within agriculture. The paper describes datasets available, performance metrics used in evaluating models, as well as implementation issues around using deep learning technology to diagnose plant disease. The paper concludes that deep learning models are effective at accurately classifying plant diseases, providing support for early diagnosis of disease, thus improving crop management and ultimately enhancing agricultural production.

**Research Gap:** Currently, most methods for monitoring crop health rely on manual monitoring, UAV imaging, or costly, labor-intensive multispectral/hyperspectral systems. In addition, many current methods focus on single-robot operation or stop-and-scan operations, which limit the ability to continuously monitor a field. Certainly, there is a need for a low-cost multi-robot ground platform that enables continuous leaf color sensing, autonomous navigation, and real-time, wireless data sharing. CropSentry is designed to fulfill this need by performing operations using two ESP32 robots operating in the same field and by implementing the TCS34725 technology to monitor leaves and the environment, creating a crop health map.

## III. Methodology

The proposed system is a multi-robot swarm coordinated approach to detect early stress in crops using multiple types of sensors. The workflow includes hardware development, calibration of sensors, autonomous navigation, wireless communication of data, visualization of data in dashboard form, and experimental validation. The complete system workflow is shown in Figure 1.

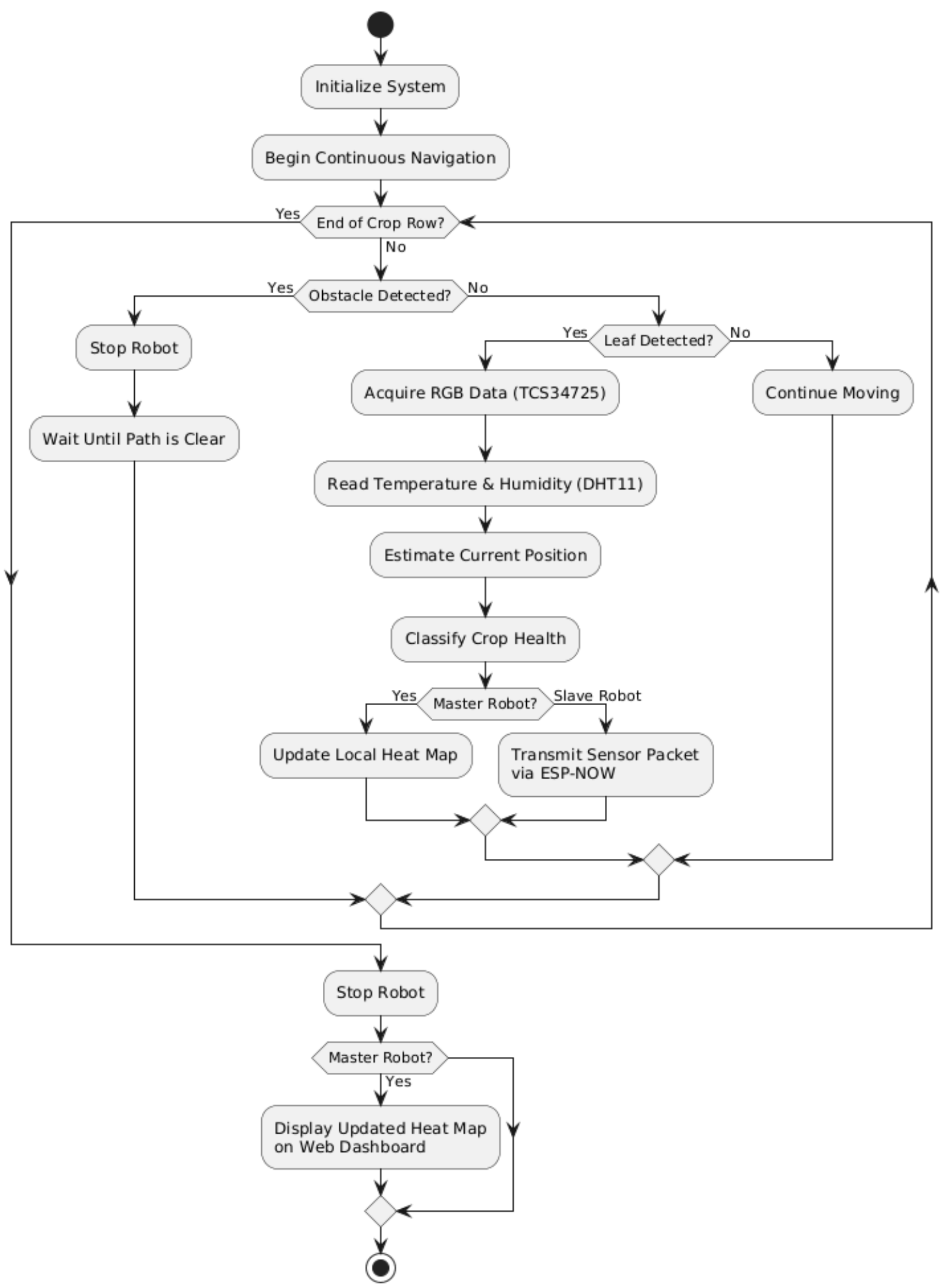

Figure 1: System Workflow

**System overview:** The system consists of two autonomous robots that run in parallel on separate agricultural crop rows. Each robot is built around an ESP32 microcontroller and is equipped with:

1. TCS34725 RGB color sensor
2. DHT11 temperature and humidity sensor
3. Front ultrasonic sensor (obstacle detection)
4. Side-mounted ultrasonic sensor (leaf detection)
5. Two continuous rotation servos for locomotion

Each slave robot collects data about the plant row and sends all the information to the master robot via ESP-NOW communication. The master robot stores all the data it has received from the Slave and contains a web-based dashboard to display and analyze that data.

**Operational Workflow:** Both robots start by initializing the complete system, establishing their internal communication line, and moving forward continuously along the rows of crops. Instead of having a stop-and-scan inspection process that other systems like to use, the robots run continuously while they are inspecting the area. While moving forward, the front ultrasonic sensor continuously captures information about the area ahead so that it is able to detect an obstacle. If there is an obstacle on its path, it will stop until the path is clear. At the same time, a side-mounted ultrasonic sensor assists in detection of the nearby leaves. Whenever there is a leaf detected, the TCS34725 sensor will collect relevant leaf color data along with ambient temperature and humidity data from the DHT11 sensor. These processes will be performed in real time, and the robot will carry on its movement without a moment's pause.

**Mechanical design:** The isometric view of the complete bot is shown in Figure 2. A convenient mechanical tank body was built in the mobile robot chassis that allows for the implementation of the required components for continuous monitoring of crop rows: sensing, processing, and moving components. The design consists of an ethical tank body with a cover used for placing the components mounts. In addition, specific wheel rim holders are used to place the locomotion assembly. The ultrasonic sensor mount for the navigation system has been fixed at the front of the robot. In the meantime, the plant ultrasonic sensor mount is at the side of the robot so that it faces the plant. As for the color sensor, it has been mounted correctly using a particular structure. The mechanical design is modular, so components can be moved or replaced independently.

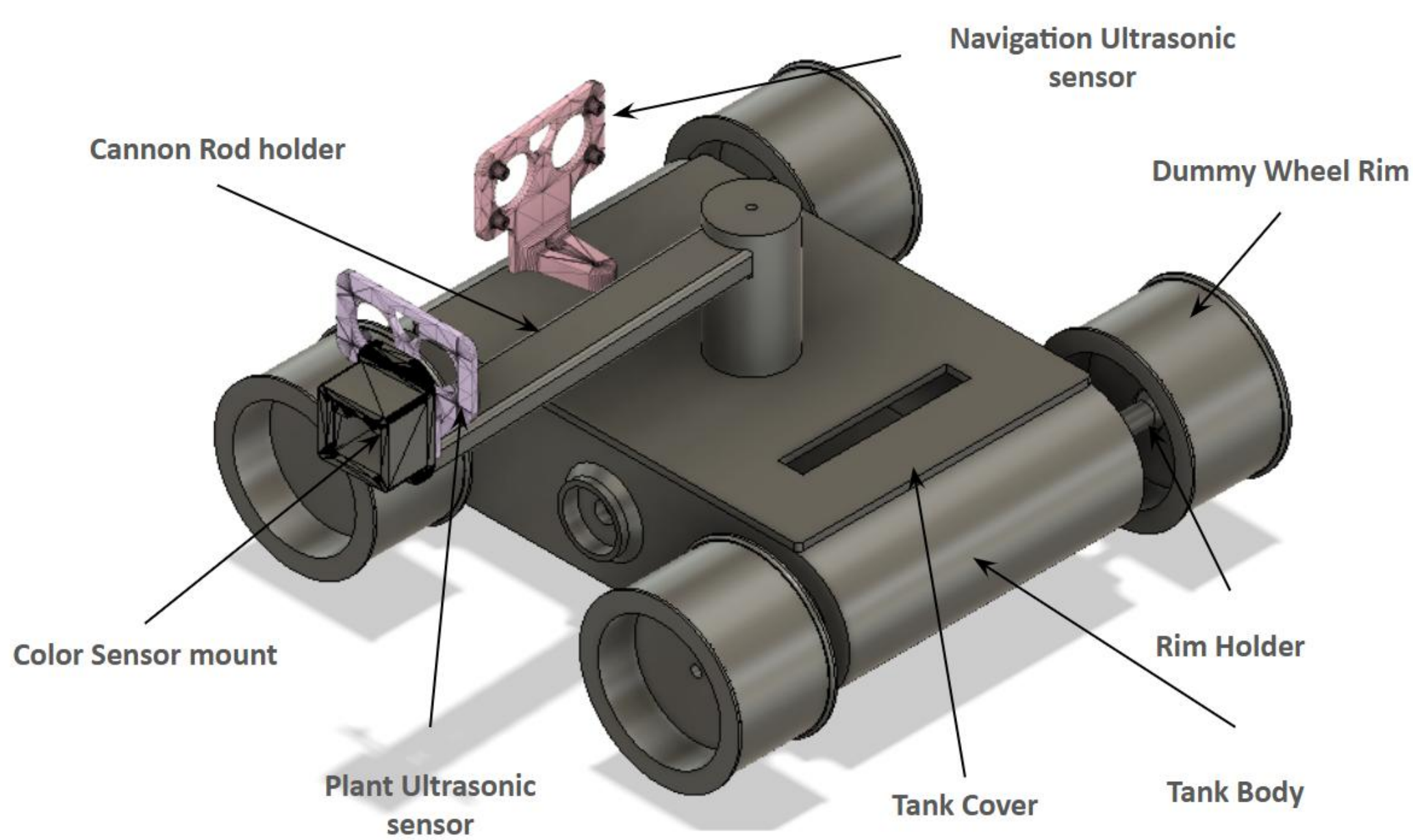

Figure 2: Isometric view of the 3D printed bot

**Hardware Implementation:** To accomplish the proposed system, an ESP32 microcontroller has been installed on each robot as its processing unit, and two continuous rotation MG996R servos provide differential drive locomotion functions. Each robot includes an obstacle detection ultrasonic sensor HC-SR04 that will prevent collisions with objects, and a second HC-SR04 ultrasonic sensor that will identify and trigger data collection of plant leaves in proximity. The complete system's circuit schematic is shown in Figure 3. The actual master and slave bot with the integration of the CAD designs and the electronics part is shown in Figure 4. The master bot is the one that intakes plant health values of both the rows and simultaneously updates them in the real-time dashboard, as shown in Figure 5.

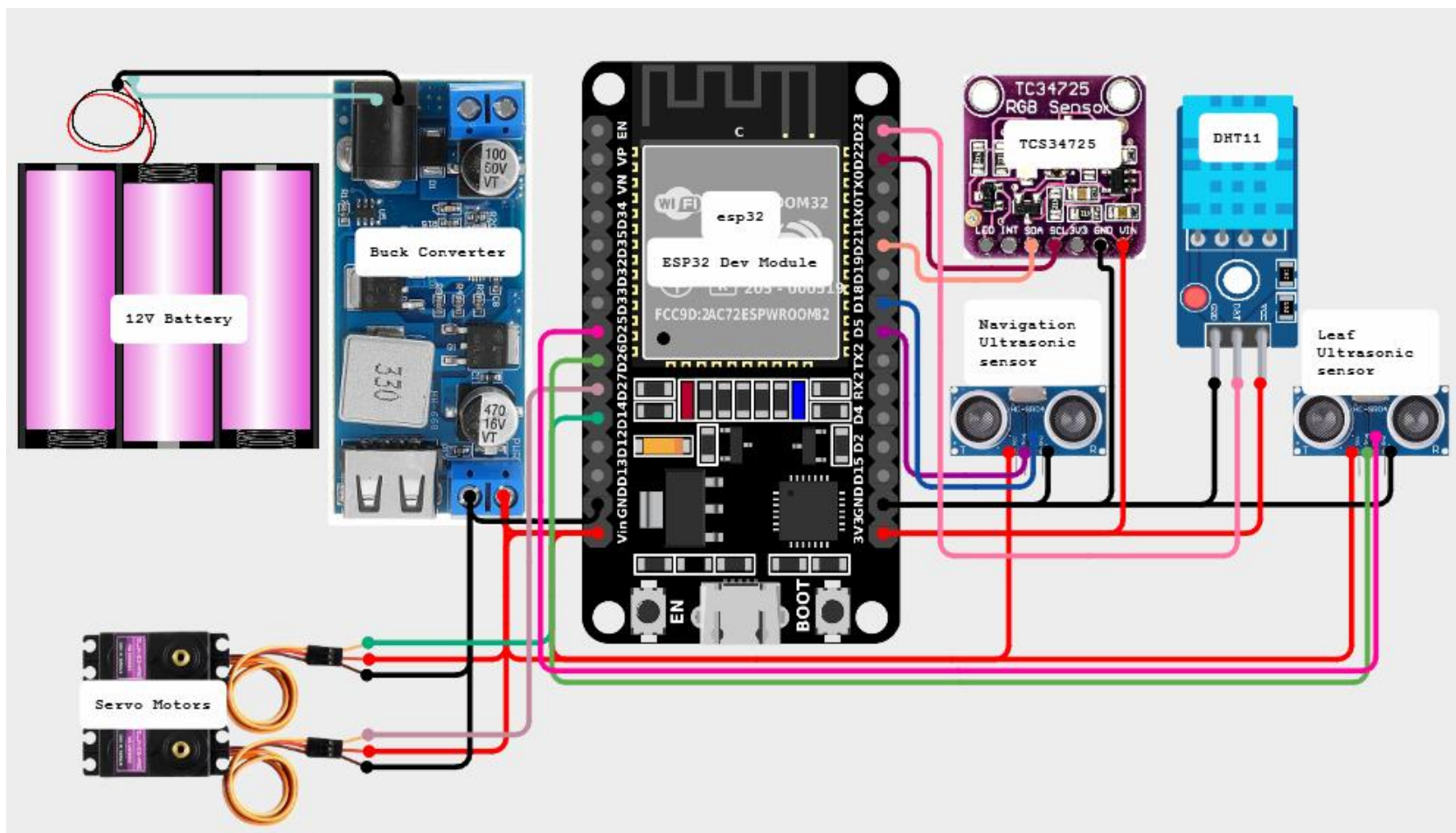

Figure 3: Circuit schematic of each bot

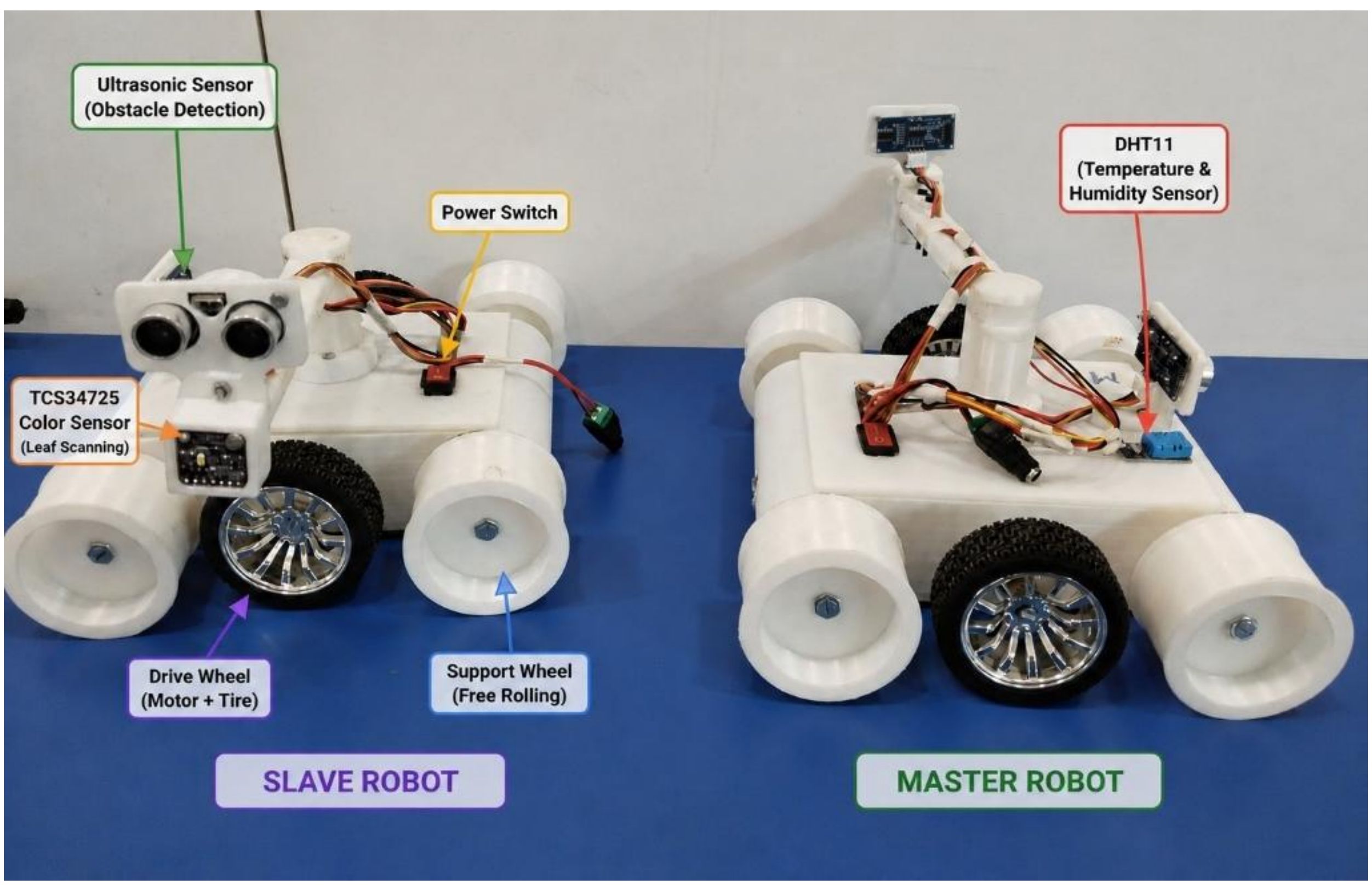

Figure 4: Master and Slave bots

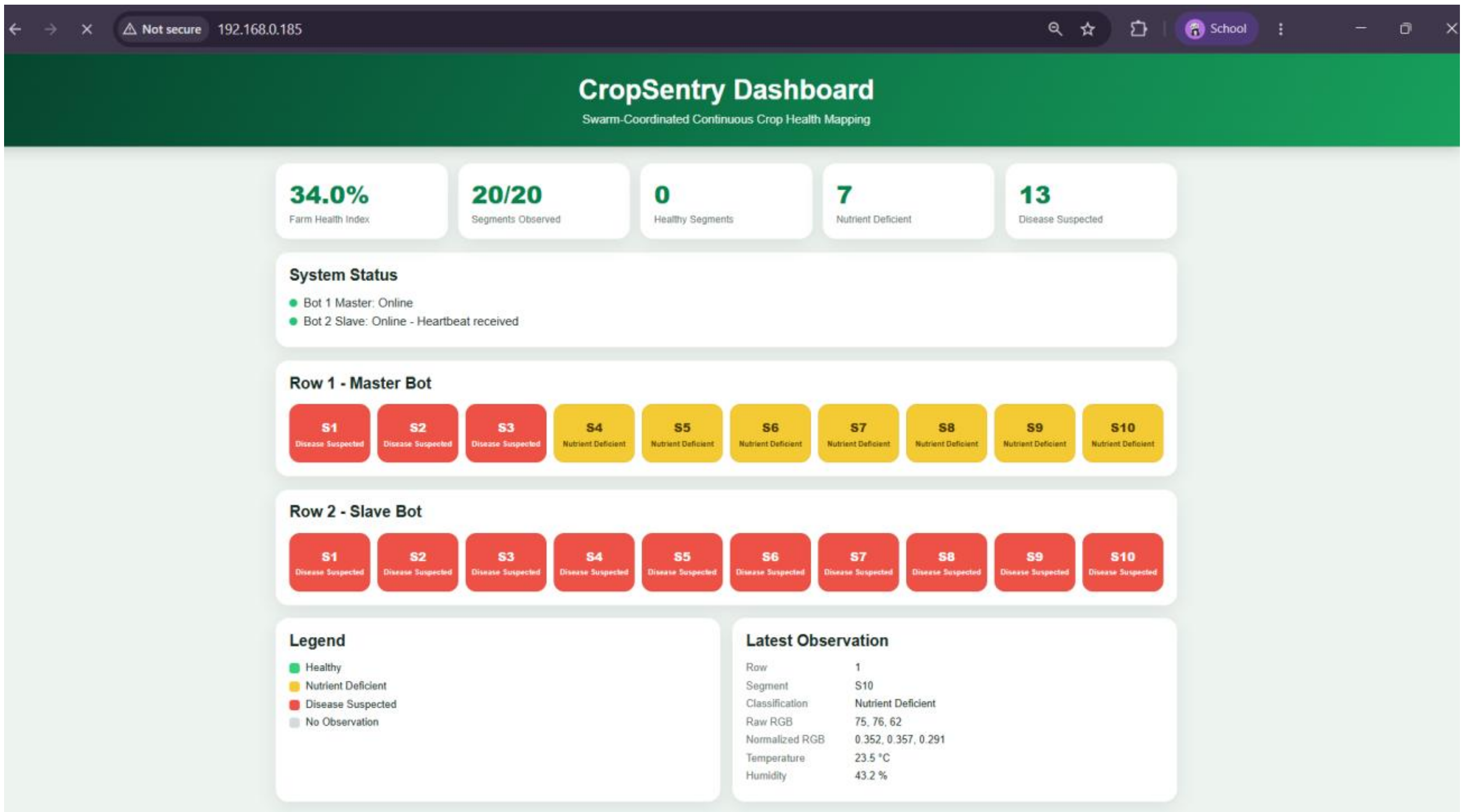

Figure 5: Real-time updating dashboard

**Sensor Calibration:**

- TCS34725 Color sensor calibration: With the help of healthy plants that presented green foliage, the calibration of the color sensor was carried out. A series of raw RGB data was obtained from different samples in order to determine the standard color parameters. To reduce the illumination intensity, normalized RGB values are obtained as mentioned in Equation 1

$$R_n = \frac{R}{R+G+B} \qquad G_n = \frac{G}{R+G+B} \qquad B_n = \frac{B}{R+G+B} \tag{1}$$

Normal conditions for vegetative plants can be identified with a normalized greenness found, but when the plant is under stress, there is a higher dominance of red with a decrease in green, which indicates that chlorophyll is degraded and not functioning properly.

Crop health can be identified through a combination of checking the color of the leaves and the condition of the environment. TCS34725 gives information about the transformation in color due to the decline of chlorophyll content and the yellowing of leaves, together with DHT11, which gives information about environmental conditions. Every observation made with the help of a TCS34725 sensor is analyzed and placed in one of the three categories. Those are:

- Healthy
- Nutrient Deficient
- Disease Suspected

- Ultrasonic sensor calibration: Two ultrasonic sensors were calibrated using a set of distances to determine the distance measurement capabilities of those sensors. The ultrasonic front detection threshold was chosen to ensure that no unexpected objects would cause a collision with the leaf insect prior to being detected. The threshold of the ultrasonic leaf detection was optimized to initiate a reading only when a leaf is present within the range of the TCS34725 sensor.

**Continuous Data Acquisition and Spatial Mapping:** The proposed system uses a continuous row sensing approach. For this method, the path taken by every crop row is divided into several segments of equal distance. Additionally, an approximate position of the robot is determined based on the distance covered.

$$D = v \times t$$

Where D = distance travelled (m), v = robot velocity (m/s) and t = elapsed time (s)
Every observation is linked to the specific row segment instead of the plant. This results in having a continuous crop health map without taking the spacing of the plants into consideration.

**Dashboard development:** The ESP-NOW communication protocol provides communication with low latency between both robots. A particular sensing event is sent as a data packet containing the following data.

- Row Index
- Spatial Segment ID
- RGB color values
- Temperature
- Humidity Level
- Health Classification Level

The slave device will transmit the observation to the master immediately after classification. The master stores both locally captured and received information in one two-dimensional spatial database $FarmMap[Row][Segment]$. This centralized structure is responsible for the real-time crop health visualization. A compact web dashboard is built with the IP of the master ESP32 and can be reached through a regular web browser with the help of the Wi-Fi network. Each spatial segment is color-coded according to the classified health condition as mentioned -

- Green - Healthy
- Yellow - Nutrient Deficient
- Red - Disease Suspected

**Experimental Evaluation:** The suggested methods are tested using beans grown under controlled environmental conditions that mimic various levels of stress. Three experimental setups are established:

- Healthy plants receiving sufficient water and nutrients.
- Drought-stressed plants with reduced irrigation.
- Nutrient-deficient plants cultivated with limited fertilizers.

The robots operate simultaneously across different rows of crops and continuously collect data from the sensors. Classification performance is quantified using Equation 2

$$Accuracy = \frac{Correct\ classification}{Total\ Observations} \times 100 \qquad (2)$$

During the experimental evaluation, various health maps are analyzed against manual visual examination to assess the precision of the stress identification process. The bots were tested in a garden with various plant types, as shown in Figure 6.

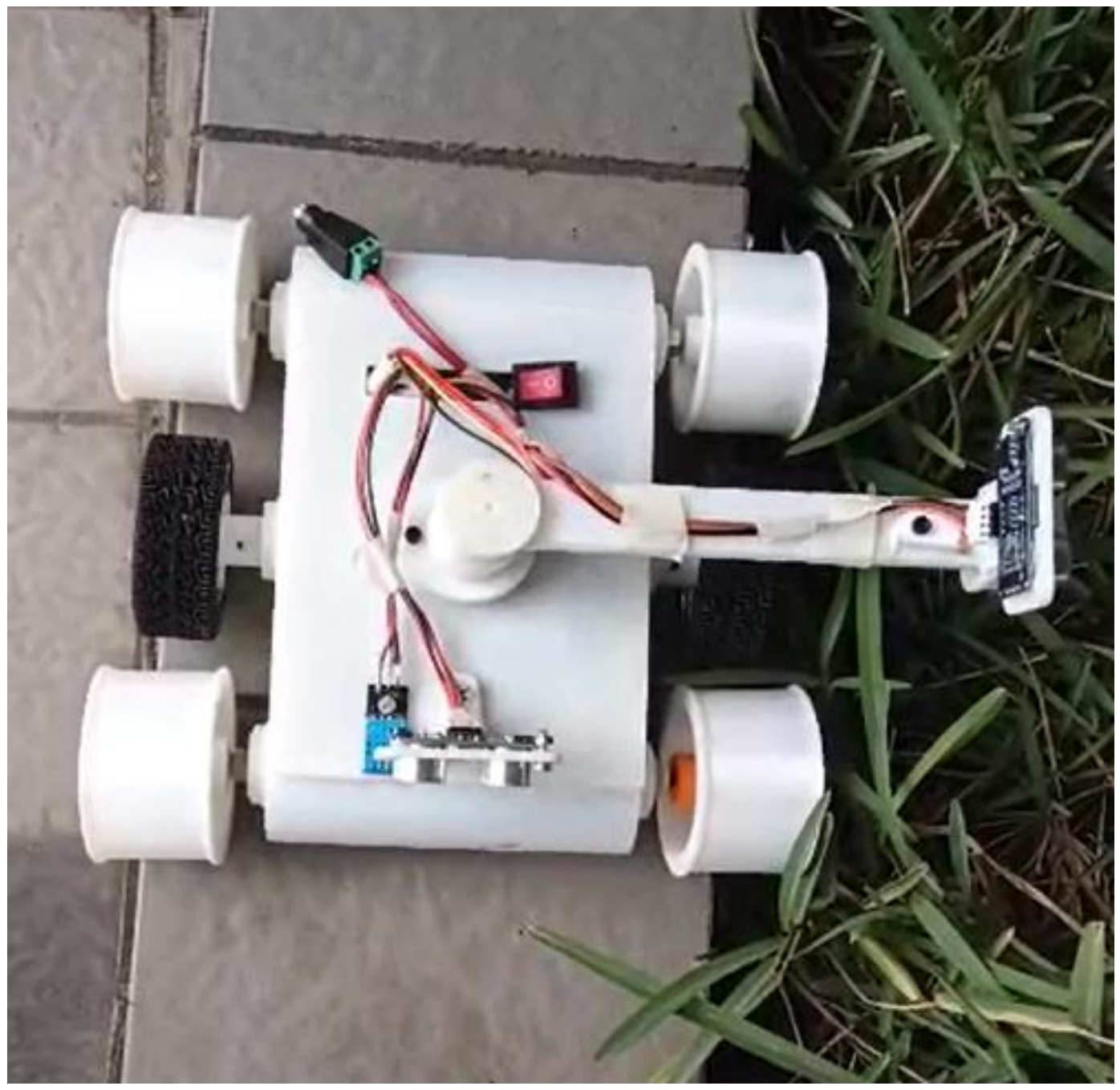
Figure 6: Testing of the bot with plants

## IV. Results and Discussions

**Overall System Performance:** The CropSentry prototype has demonstrated its ability to integrate sophisticated technologies, including autonomous ground navigation, leaf detection, crop-health classification based on RGB monitoring, environmental sensing, wireless communication, and centralized spatial visualization. Two robots made on the basis of ESP32 were programmed to work in different lines of agriculture, with the Master robot responsible for in situ sensing and running of the dashboard, while the Slave robot sent its data to the Master with the help of ESP-NOW. The technological implementation made it possible to work with 10 spatial segments in each row and conduct mapping in 20 spatial segments over two lines of agriculture. The robots were on the move all the time while operating normally. The first ultrasonic sensor was equipped with a 20 cm obstacle detection threshold, and the other sensor was equipped with a 25 cm detection threshold for leaf scanning. A scan interval of at least 1200 ms was assumed in order to avoid periodic measurements of the same leaf. Table 1 shows the overall performance of the prototype, including parameter settings and success rate.

Table 1: Overall Prototype Performance

| Parameter | Experimental Result |
|---|---|
| Number of robots | 2 |
| Number of crop rows monitored simultaneously | 2 |
| Health categories | 3 |
| Color sensor | TCS34725 |
| Environmental sensor | DHT11 |
| Wireless communication | ESP-NOW |
| Navigation mode | Continuous |
| few Dashboard | Real-time centralized |
| Successful robot operation | [06/08 trials] |
| Successful wireless communication | [03/03 observations] |

The success of the synchronous functioning of both robots indicates that economically distributed robotic architecture can be effectively applied for agricultural monitoring purposes. Unlike a single sensing unit, whereby only one sensing unit would be used to monitor one crop row, here several monitoring units can monitor two crop rows simultaneously, thus maximizing the area covered by monitoring with not much increase in terms of monitoring challenges for each unit.

**Leaf-Color Classification Performance:** TCS34725 measurements provided critical data to identify leaf shades differently. The normalized values of red, green, and blue were more useful for classification than real RGB values because the results depended on how significant each of the RGB values was for classification. In general, non-diseased leaves had a more pronounced green color, which was observed through higher normalized green values. Yellow leaves were distinguished by low green values, indicating a high similarity between the red and green components in the monitored signal. The samples that were considered to have some diseases were characterized neither by the features of healthy leaves nor by the features of leaves that lack nutrients. The overall classification accuracy was calculated using Equation 2 and presented in Table 2.

Table 2: Crop-Health Classification Results

| **Actual Condition** | **Number Tested** | **Correctly Classified** | **Incorrectly Classified** | **Accuracy** |
|---|---|---|---|---|
| Healthy | 23 | 19 | 4 | 82.60% |
| Nutrient Deficient | 25 | 22 | 3 | 88% |
| Disease Suspected | 15 | 12 | 3 | 80% |
| Overall | 63 | 53 | 10 | 84.12% |

**Analysis of misclassification:** The main reasons for classification errors can be attributed to the similarity in color properties of different crop stress states. Since TCS34725 measures the reflected RGB elements on the leaf surface, the output is influenced not only by the plant's health condition but also by the light intensity, leaf position, leaf surface properties, and other biological variability factors. In general, healthy vegetation has a stronger green response and is, to some extent, easier to identify. In general, nutrient-deficient leaves may lose green pigmentation gradually and will increase yellow coloration. Affected by the disease, leaves may also develop yellow, brown, or very light areas, making them resemble leaves with a nutrient deficiency. As a result, a part of the observations can be classified into the neighboring class. The classification errors are particularly relevant for early stress detection, when the visible symptoms can be weak and have not yet formed a specific color pattern.

**Continuous Robot Navigation and Detection:** One significant effect of the experimental assessment was that sensor readings could successfully be transformed into a constantly updated representation of the state of a crop. Moreover, the dashboard presented a visualization of data from the two robots used, so that the operator could locate regions with healthy vegetation and spots that needed monitoring.  Another positive aspect is that the use of the robots being in a constant motion does not require the robot to stop at each individual plant. Thus, it provides for better operational efficiency and enables the system to process a larger part of the crop row. The created visualization turned out to be very helpful when identifying the presence of the local stress zones. The dashboard showed the operator much more than simple figures obtained through RGB scans, as it transformed the decision-making results into a color-coded visualization. The updated dashboard is shown in Figure 7.

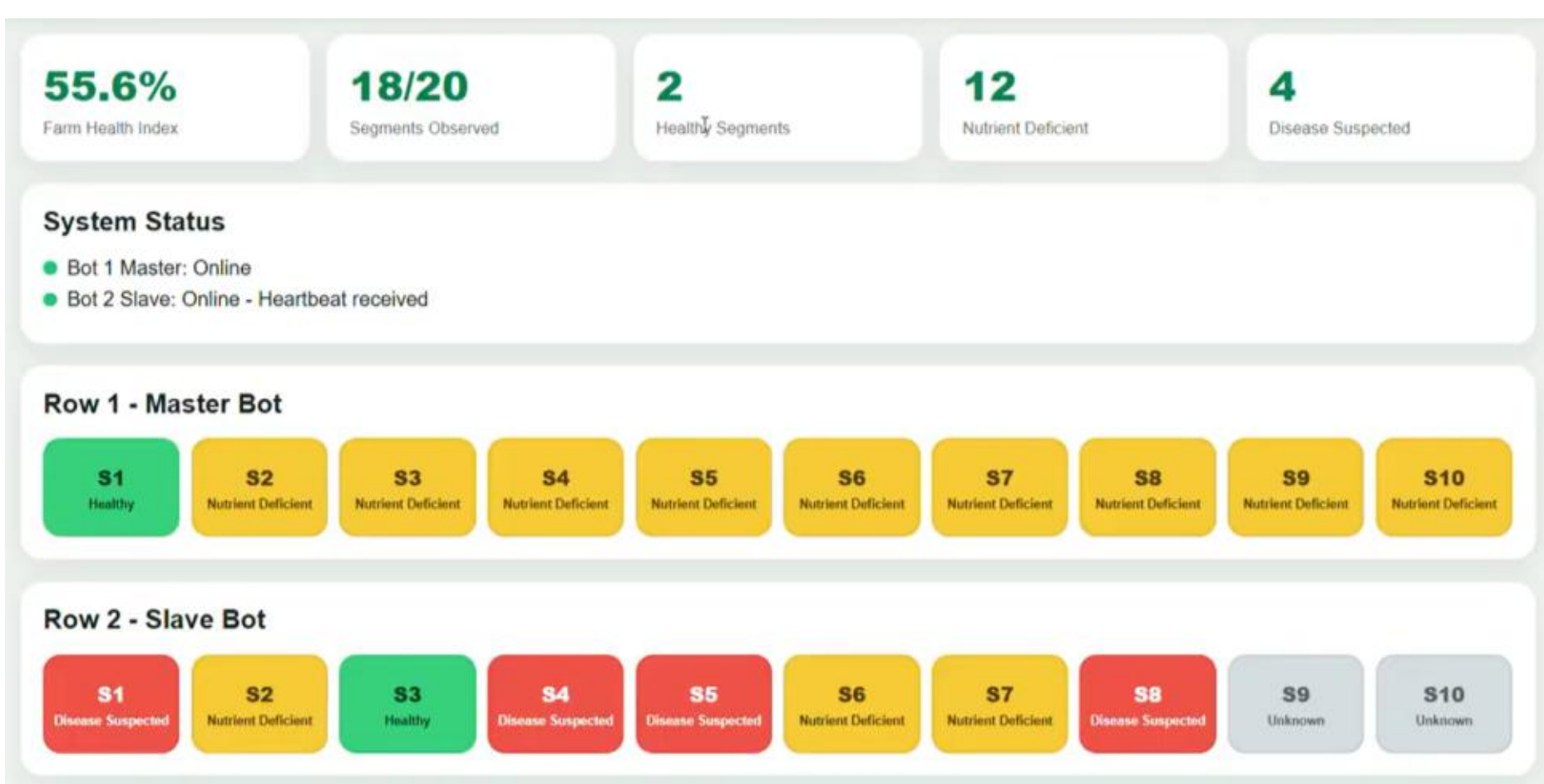

Figure 7: Health classifying dashboard visualization

**Multi-Robot Communication Performance:** The experimental procedure proved that coordination between the robots could be achieved using data communication. The Slave transmitted its measurement data to the Master, which then integrated that data with its own. By employing ESP-NOW, data was transferred between the two ESP32 devices while eliminating the need for the Slave to keep an independent dashboard active. The result was a centralized monitoring system in which the Master served as both a data collection station and a visualization system, as presented in Table 3.

Table 3: Wireless Communication Performance

| Parameter | Result |
|---|---|
| Total Slave observations generated | 10 |
| Successfully received by the Master | 10 |
| Lost observations | 0 |
| Communication success rate | 100% |
| Duplicate packets detected | 0 |
| Communication failures | 0 |

**Environmental Measurements:** The DHT11 provided temperature and humidity measurements simultaneously with leaf-color observations. These measurements provided an environmental context for interpreting changes in leaf appearance. The measurements are as mentioned in Table 4. Environmental measurements cannot function on their own as diagnostic indicators of a particular disease. However, they provide valuable context when seeking to understand the variation in plant appearance during the field operation. For instance, variations in humidity and temperature affect leaf state and thus influence RGB readings.

Table 4: Environmental Conditions During Testing

| Parameter | Minimum | Maximum |
|---|---|---|
| Temperature (°C) | 27 | 35 |
| Relative Humidity (%) | 55 | 69 |

## V. Conclusion

The successfully created prototype presented a cost-effective, coordinated way to monitor crops using two autonomous robots based on ESP32 microcontrollers. This system managed to observe 2 rows of crops simultaneously and perform RGB and environmental sensing, as well as ultrasonic observations, identifying crop health levels in three categories: Healthy, Nutrient Deficient, or Disease Suspected. Unlike other management systems that relied on the stop-and-scan method, the robots performed uninterrupted monitoring of crop health because signals were sent and received via ESP-NOW from the Slave robot to the Master robot, which then combined all data. In addition, the Master was able to produce a real-time spatial map of crop health status using a dashboard that supported observations of 20 segments, with 10 in one row. The crop-sensing system appeared to be very good, as it employed thresholds of 20 cm for object detection, 25 cm for leaf detection, and 2 seconds for refreshing the dashboard. The system provides an affordable alternative for preliminary crop-stress detection and spatial identification of potentially affected regions. However, RGB-based classification remains sensitive to illumination, leaf orientation, and overlap between nutrient-deficiency and disease symptoms. Future work will focus on larger field trials, improved illumination compensation, wheel-based odometry for more accurate spatial localization, and integration of additional spectral or soil-moisture sensing to improve classification reliability and scalability.

## Acknowledgements

I sincerely thank all my mentors for their continuous guidance, technical support, and valuable feedback throughout the development of this project.